\pdfoutput=1

\documentclass[11pt]{article}

\usepackage[final]{emnlp2021}

\usepackage{times}
\usepackage{latexsym}

\usepackage{tikz}
\usetikzlibrary{fit, quotes, matrix}
\usepackage{subfigure}
\usepackage{multirow}
\usepackage{booktabs}
\usepackage{amsthm,amsmath,amssymb}
\usepackage{mathrsfs}
\usepackage{algorithm}
\usepackage{array}
\usepackage{algpseudocode}
\usepackage{pifont}
\usepackage{pgfplots}
\usepackage[T1]{fontenc}
\usepackage[utf8]{inputenc}

\usepackage{microtype}

\title{Honey or Poison? Solving the Trigger Curse in Few-shot Event Detection via Causal Intervention}

\author{
  Jiawei Chen${}^{1,3}$,
  Hongyu Lin${}^{1,}$\footnotemark[1],
  Xianpei Han${}^{1,2,}$\footnotemark[1]\thanks{~ Corresponding authors.},
  Le Sun${}^{1,2}$,
  \\
  ${}^{1}$Chinese Information Processing Laboratory ~
  ${}^{2}$State Key Laboratory of Computer Science \\
  Institute of Software, Chinese Academy of Sciences, Beijing, China\\
  ${}^{3}$University of Chinese Academy of Sciences, Beijing, China \\
 {\tt \{jiawei2020\}@iscas.ac.cn} \\
 {\tt \{hongyu,xianpei,sunle\}@iscas.ac.cn} \\
}

\begin{document}
  \maketitle
  \begin{abstract}
    Event detection has long been troubled by the \emph{trigger curse}: overfitting the trigger will harm the generalization ability while underfitting it will hurt the detection performance. This problem is even more severe in few-shot scenario. In this paper, we identify and solve the trigger curse problem in few-shot event detection (FSED) from a causal view. By formulating FSED with a structural causal model (SCM), we found that the trigger is a confounder of the context and the result, which makes previous FSED methods much easier to overfit triggers. To resolve this problem, we propose to intervene on the context via backdoor adjustment during training. Experiments show that our method significantly improves the FSED on ACE05, MAVEN and KBP17 datasets.  
  \end{abstract}
  \tikzset{
    blue/.style = {
      shape = rectangle,
      draw = {rgb,255:red,50; green,118; blue,179},
      text width = 5.5cm,
      align = center,
      rounded corners = 2mm,
      minimum width = 1cm,
      minimum height = 0.75cm,
      fill = {rgb,255:red,222; green,235; blue,246},
    },
    yellow/.style = {
      shape = rectangle,
      draw = {rgb,255:red,254; green,216; blue,111},
      text width = 3cm,
      rounded corners = 2mm,
      minimum width = 1cm,
      minimum height = 0.75cm,
      fill = {rgb,255:red,255; green,241; blue,206},
    },
    green/.style = {
      shape = rectangle,
      draw = {rgb,255:red,84; green,128; blue,57},
      text width = 4cm,
      rounded corners = 2mm,
      minimum width = 1cm,
      minimum height = 0.75cm,
      fill = {rgb,255:red,226; green,240; blue,218},
    },
    orange/.style = {
      shape = rectangle,
      draw = black,
      dashed,
      text width = 3cm,
      rounded corners = 2mm,
      minimum width = 1cm,
      minimum height = 1 cm,
      fill = {rgb,255:red,250; green,229; blue,215},
    },
    arrow/.style = {
      -latex,
      draw = black,
      thick,
    },
    rectangle/.style={
        shape = circle,
        draw = black,
        text width = 0.2cm,
        align = center,
        minimum height = 0.2cm,
        thick,
    },
    bluerectangle/.style={
        shape = circle,
        draw = {rgb,255:red,50; green,118; blue,179},
        text width = 0.2cm,
        align = center,
        minimum height = 0.2cm,
        thick,
        fill = {rgb,255:red,222; green,235; blue,246},
    },
    orangerectangle/.style={
        shape = circle,
        draw = {rgb,255:red,195; green,90; blue,32},
        text width = 0.2cm,
        align = center,
        minimum height = 0.2cm,
        thick,
        fill = {rgb,255:red,250; green,229; blue,215},
    },
    greenrectangle/.style={
        shape = circle,
        draw = {rgb,255:red,84; green,128; blue,57},
        text width = 0.2cm,
        align = center,
        minimum height = 0.2cm,
        thick,
        fill = {rgb,255:red,226; green,240; blue,218},
    },
    yellowrectangle/.style={
        shape = circle,
        draw = {rgb,255:red,254; green,216; blue,111},
        text width = 0.2cm,
        align = center,
        minimum height = 0.2cm,
        thick,
        fill = {rgb,255:red,255; green,241; blue,206},
    },
  }

  \section{Introduction}
  
  Event detection (ED) aims to identify and classify event triggers in a sentence, e.g., detecting an \texttt{Attack} event triggered by \emph{fire} in ``They killed by hostile fire in Iraqi''. Recently, supervised ED approaches have achieved promising performance \citep{chen-etal-2015-event,nguyen-grishman-2015-event,nguyen-etal-2016-joint,lin-etal-2018-nugget,lin-etal-2019-sequence,lin-etal-2019-cost,du-cardie-2020-event,liu-etal-2020-mrc,lu-etal-2021-text2event}, but when adapting to new event types and domains, a large number of manually annotated event data is required which is expensive. By contrast, few-shot event detection (FSED) aims to build effective event detectors that are able to detect new events from instances (query) with a few labeled instances (support set). Due to their ability to classify novel types, many few-shot algorithms have been used in FSED, e.g., metric-based methods like Prototypical Network \citep{lai-etal-2020-extensively,deng2020meta,cong-etal-2021-shot}.
  
  \begin{figure}
    \setlength{\abovecaptionskip}{0.2cm}
    \setlength{\belowcaptionskip}{-0.7cm}
    \begin{center}
      \subfigure[Structural Causal Model for the data distribution of FSED]{
        \label{fig:scm}
        \resizebox{0.45\textwidth}{!}{\begin{tikzpicture}
        \node[blue,text width = 2.2cm] (e)  at(-7,2) {An \texttt{Attack} event in Iraqi.};
        \node[blue,text width = 1cm,align=center] (t)  at(-3.5,2) {\emph{\textbf{fire}}};
        \node[blue,text width = 4.0cm,align=center] (c)  at(-5.5,-0.6) {They were killed by hostile \textbf{[MASK]} in Iraqi.};
        \node[blue,text width = 3.5cm,align=center] (s)  at(-0.5,-0.6) {They were killed by hostile \emph{\textbf{fire}} in Iraqi.};
        \node[green,text width = 1.5cm,align=center] (y)  at(3.5,-0.6) {1 or 0};
        \node[orange,text width = 1cm,align=center] (q)  at(5,1) {\textcolor[RGB]{195,90,32}{\large \emph{\textbf{Query}}}};
        \node at(-7,2.9) {$E$};
        \node at(-3.5,2.7) {$T$};
        \node at(-5.5,-1.5) {$C$};
        \node at(-0.5,-1.5) {$S$};
        \node at(3.5,-1.5) {$Y$};
        \node[bluerectangle] (e2)  at(-1,2) {\small $E$};
        \node[bluerectangle] (c2)  at(-0.1,0.9) {\small$C$};
        \node[bluerectangle] (t2)  at(0.8,2) {\small$T$};
        \node[bluerectangle] (s2)  at(1.7,0.9) {\small$S$};
        \node[greenrectangle] (y2)  at(3,0.9) {\small$Y$};
        \node[orangerectangle] (q2)  at(3.7,2) {\small$Q$};
        \draw[arrow] (e2) -- (t2);
        \draw[arrow] (e2) -- (c2);
        \draw[arrow] (t2) -- (s2);
        \draw[arrow] (t2) -- (c2);
        \draw[arrow] (c2) -- (s2);
        \draw[arrow] (s2) -- (y2);
        \draw[arrow] (q2) -- (y2);
        \draw[arrow] (-6.9,1.4) -- (-5.4,0.1);
        \draw[arrow] (-3.7,1.4) -- (-5.2,0.1);
        \draw[arrow] (-3.3,1.4) -- (-1.3,0.1);
        \draw[arrow] (c) -- (s);
        \draw[arrow] (s) -- (y);
        \draw[arrow] (q) -- (y);
        \draw[arrow] (e) -- (t);
        \draw[thick, dashed] (-1.5,2.5) -- (4.2,2.5) -- (4.2,0.5) -- (-1.5,0.5) -- (-1.5,2.5);
        \end{tikzpicture}}
      }
      \subfigure[The data distribution of FSED after causal intervention]{
        \label{fig:scmdos}
        \resizebox{0.45\textwidth}{!}{\begin{tikzpicture}
        \node[blue,text width = 2.2cm] (e)  at(-7,2) {An \texttt{Attack} event in Iraqi.};
        \node[yellow,align=center,text width = 2.5cm] (t)  at(-3.5,2) {\emph{\textbf{forces, fire, attack, ...}}};
        \node[blue,text width = 4.0cm,align=center] (c)  at(-5.5,-0.6) {They were killed by hostile \textbf{[MASK]} in Iraqi.};
        \node[yellow,text width = 3.5cm,align=center] (s)  at(-0.5,-0.6) {They were killed by hostile \textbf{[T]} in Iraqi.};
        \node[green,text width = 1.5cm,align=center] (y)  at(3.5,-0.6) {1 or 0};
        \node[orange,text width = 1cm,align=center] (q)  at(5,1) {\textcolor[RGB]{195,90,32}{\large \emph{\textbf{Query}}}};
        \node at(-7,1.1) {$E$};
        \node at(-3.5,1.1) {$T$};
        \node at(-5.5,-1.5) {$C$};
        \node at(-0.5,-1.5) {$S$};
        \node at(3.5,-1.5) {$Y$};
        \node[bluerectangle] (e2)  at(-1,2) {\small $E$};
        \node[bluerectangle] (c2)  at(-0.1,0.9) {\small$C$};
        \node[yellowrectangle] (t2)  at(0.8,2) {\small$T$};
        \node[yellowrectangle] (s2)  at(1.7,0.9) {\small$S$};
        \node[greenrectangle] (y2)  at(3,0.9) {\small$Y$};
        \node[orangerectangle] (q2)  at(3.7,2) {\small$Q$};
        \draw[arrow] (e2) -- (t2);
        \draw[arrow] (t2) -- (s2);
        \draw[arrow] (c2) -- (s2);
        \draw[arrow] (s2) -- (y2);
        \draw[arrow] (q2) -- (y2);
        \draw[arrow] (-3.1,1.2) -- (-1.3,0.1);
        \draw[arrow] (c) -- (s);
        \draw[arrow] (s) -- (y);
        \draw[arrow] (q) -- (y);
        \draw[arrow] (e) -- (t);
        \draw[thick, dashed] (-1.5,2.5) -- (4.2,2.5) -- (4.2,0.5) -- (-1.5,0.5) -- (-1.5,2.5);
        \end{tikzpicture}}
      }
      \caption{Illustration of the causal intervention strategy proposed in this paper. The graph includes the event $E$, the trigger set $T$, the context set $C$, the support instance $S$, the prediction $Y$ and the query instance $Q$.}
    \end{center}
  \end{figure}
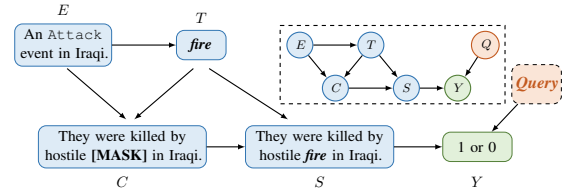
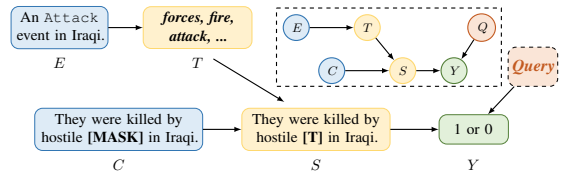
  
  Unfortunately, there has long been a “trigger curse” which troubles the learning of event detection models, especially in few-shot scenario \citep{bronstein-etal-2015-seed,liu-etal-2017-exploiting,chen-etal-2018-collective,liu2019exploiting,ji-etal-2019-exploiting}. For many event types, their triggers are dominated by several popular words, e.g., the \texttt{Attack} event type is dominated by \emph{war, attack, fight, fire, bomb} in ACE05. And we found the top 5 triggers of each event type cover 78\% of event occurrences in ACE05. Due to the trigger curse, event detection models nearly degenerate to a trigger matcher, ignore the majority of contextual information and mainly rely on whether the candidate word matches the dominant triggers. This problem is more severe in FSED: since the given support instances are very sparse and lack diversity, it is much easier to overfit the trigger of the support instances. An intuitive solution for the trigger curse is to erase the trigger information in instances and forces the model to focus more on the context. Unfortunately, due to the decisive role of triggers, directly wiping out the trigger information commonly hurts the performance \citep{lu-etal-2019-distilling,liu-etal-2020-context}. Some previous approaches try to tackle this problem by introducing more diversified context information like event argument information \citep{liu-etal-2017-exploiting,liu2019exploiting,ji-etal-2019-exploiting} and document-level information \citep{ji-grishman-2008-refining,liao-grishman-2010-using, duan2017exploiting, chen-etal-2018-collective}. However, rich context information is commonly not available for FSED, and therefore these methods can not be directly applied.

  In this paper, we revisit the trigger curse in FSED from a causal view. Specifically, we formulate the data distribution of FSED using a trigger-centric structural causal model (SCM) \citep{pearl2016causal} shown in Figure \ref{fig:scm}. Such trigger-centric formulation is based on the fact that, given the event type, contexts have a much lower impact on triggers, compared with the impact of triggers on contexts. This results in the decisive role of triggers in event extraction, and therefore conventional event extraction approaches commonly follow the trigger-centric procedure (i.e., identifying triggers first and then using triggers as an indicator to find arguments in contexts). Furthermore, the case grammar theory in linguistics \cite{fillmore1967case} also formulate the language using such trigger/predicate-centric assumption, and have been widely exploited in many NLP tasks like semantic role labeling \cite{gildea2002automatic} and abstract meaning representation \cite{banarescu-etal-2013-abstract}.

  From the SCM, we found that $T$(trigger set) is a confounder of the $C$(context set) and the $Y$(result), and therefore there exists a backdoor path $C\leftarrow T \to Y$. The backdoor path explains why previous FSED models disregard contextual information: it misleads the conventional learning procedure to mistakenly regard effects of triggers as the effects of contexts. Consequently, the learning criteria of conventional FSED methods are optimized towards spurious correlation, rather than capturing causality between $C$ and $Y$. To address this issue, we propose to intervene on context to block the information from trigger to context. Specifically, we apply backdoor adjustment to estimate the interventional distribution that is used for optimizing causality. Furthermore, because backdoor adjustment relies on the unknown prior confounder (trigger) distribution, we also propose to estimate it based on contextualized word prediction.
  
  We conducted experiments on ACE05\footnote{https://catalog.ldc.upenn.edu/LDC2006T06}, MAVEN\footnote{https://github.com/THU-KEG/MAVEN-dataset} and KBP17\footnote{https://tac.nist.gov/2017/KBP/data.html} datasets. Experiments show that causal intervention can significantly alleviate trigger curse, and therefore the proposed method significantly outperforms previous FSED methods.
  
  \section{Structural Causal Model for FSED}
  \label{sec:model-scm}
  This section describes the structural causal model (SCM) for FSED, illustrated in Figure \ref{fig:scm}. Note that, we omit the causal structure of the query for simplicity since it is the same as the support set. Concretely, the SCM formulates the data distribution of FSED: 1) Starting from an event $E$ we want to describe (in Figure \ref{fig:scm} is an \texttt{Attack} in Iraqi). 2) The path $E\to T$ indicates the trigger decision process, i.e., selecting words or phrases (in Figure \ref{fig:scm} is \emph{fire}) which can almost clearly express the event occurrence \citep{doddington2004automatic}. 3) The path $E\to C\leftarrow T$ indicates that a set of contexts are generated depending on both the event and the trigger, which provides background information and organizes this information depending on the trigger. For instance, the context ``They killed by hostile [\emph{fire}] in Iraqi'' provides the place, the role and the consequences of the event, and this information is organized following the structure determined by \emph{fire}. 4) an event instance is generated by combining one of the contexts in $C$ and one of the triggers in $T$ via the path $C\to S \leftarrow T$. 5) Finally, a matching between query and support set is generated through $S\to Y \leftarrow Q$.

  Conventional learning criteria for FSED directly optimize towards the conditional distribution $P(Y|S,Q)$. However, from the SCM, we found that the backdoor path $C\leftarrow T\to Y$ pass on associations \citep{pearl2016causal} and mislead the learning with spurious correlation. Consequently, the learning procedure towards $P(Y|S,Q)$ will mistakenly regard the effects of triggers as the effects of contexts, and therefore overfit the trigger information.
  
  \section{Causal Intervention for Trigger Curse}\label{sec:model-ba}
  Based on the SCM, this section describes how to resolve the trigger curse via causal intervention. 
  
  \noindent\textbf{Context Intervention.} To block the backdoor path, we intervene on the context $C$ and the new context-intervened SCM is shown in Figure \ref{fig:scmdos}. Given support set $s$, event set $e$ of $s$, context set $\mathcal{C}$ of $s$ and query instance $q$, we optimize the interventional distribution $P(Y|do(C=\mathcal{C}),E=e,Q=q)$ rather than $P(Y|S=s,Q=q)$, where $do(\cdot)$ denotes causal intervention operation. By intervening, the learning objective of models changes from optimizing correlation to optimizing causality.

  \noindent\textbf{Backdoor Adjustment.} Backdoor adjustment is used to estimate the interventional distribution\footnote{The proof is shown in Appendix}:
  \begin{equation}
    \label{eq:cp}
    \small
    \begin{aligned}
       &P(Y|do(C{=}\mathcal{C}),E{=}e,Q{=}q)\\
      =&\sum\limits_{t\in T}\sum\limits_{s\in S} P(Y|s,q)P(s|\mathcal{C},t)P(t|e),
    \end{aligned}
  \end{equation}
  where $P(s|\mathcal{C},t)$ denotes the generation of $s$ from the trigger and contexts. $P(s|\mathcal{C},t)=1/|\mathcal{C}|$ if and only if the context of $s$ in $\mathcal{C}$ and the trigger of $s$ is $t$. 
  $P(Y|s,q) \propto \phi(s,q;\theta)$ is the matching model between $q$ and $s$ parametrized by $\theta$. 
  
  \noindent\textbf{Estimating $P(t|e)$ via Contextualized Prediction.} The confounder distribution $P(t|e)$ is unknown because $E$ is a hidden variable. Since the event argument information is contained in $\mathcal{C}$, we argue that $P(t|e)\propto M(t|\mathcal{C})$ where $M(\cdot|\mathcal{C})$ indicates a masked token prediction task \citep{taylor1953cloze} which is constructed by masking triggers in the support set. In this paper, we use masked language model to calculate $P(t|e)$ by first generating a set of candidate triggers through the context: $T_c=\{t_i|i=1,2,\ldots \} \cup  \{t_0\}$, where $t_i$ is the i-th predicted token and $t_0$ is the original trigger of the support set instance, then $P(t|e)$ is estimated by averaging logit obtained from the MLM:
  \begin{equation}
    \small
  P(t_i|e) = \left\{
    \begin{aligned}
      &\lambda\quad &i = 0\\
      &(1-\lambda)\frac{\exp (l_i)}{\sum_{j}\exp (l_j)}\quad &i \neq 0
    \end{aligned}
    \right.
  \end{equation}
  where $l_i$ is the logit for the $i^{th}$ token. To reduce the noise introduced by MLM, we assign an additional hyperparameter $\lambda\in (0,1)$ to $t_0$. 

  \noindent\textbf{Optimizing via Representation Learning.} 
  Given the interventional distribution, FSED model can be learned by minimizing the loss function on it:
  \begin{equation}
    \label{eq:loss_ori}
    \small
    \begin{aligned}
    \mathcal{L(\theta)} &= -\sum_{q \in Q}  f( P(Y| do(\mathcal{C}),e,q;\theta) ) \\
    &= -\sum_{q \in Q}  f(\sum\limits_{t\in T}\sum\limits_{s\in S}P(Y|s,q;\theta)P(s|\mathcal{C},t)P(t|e) )
    \end{aligned}
  \end{equation}
  where $Q$ is training queries and $f$ is a strict monotonically increasing function. However, the optimization of $\mathcal{L(\theta)}$ needs to calculate every $P(Y|s,q;\theta)$, which is quite time-consuming. To this end, we propose a surrogate learning criteria $\mathcal{L}_{SG}(\theta)$ to optimize the causal relation based on representation learning:
  
  \vspace{-0.5cm}
  \begin{equation}
    \label{eq:loss_new}
    \small
    \begin{aligned}
    \mathcal{L}_{SG}(\theta) = -\sum_{q \in Q}  g( & \mathbf{R}(q;\theta), \\
    &\sum\limits_{t\in T}\sum\limits_{s\in S} P(s|\mathcal{C},t)P(t|e) \mathbf{R}(s;\theta))
  \end{aligned}
  \end{equation}
  Here $\mathbf{R}$ is a representation model which inputs $s$ or $q$ and outputs a dense representation. $g(\cdot,\cdot)$ is a distance metric measuring the similarity between two representations. Such loss function is widely used in many metric-based methods (e.g., Prototypical Networks and Relation Networks). In the Appendix, we prove $\mathcal{L}_{SG}(\theta)$ is equivalent to $\mathcal{L(\theta)}$.

  \begin{table*}
    \tiny\addtolength{\tabcolsep}{-4pt}
    \setlength{\abovecaptionskip}{0.1cm}
    \setlength{\belowcaptionskip}{-0.4cm}
    \begin{center}
    \renewcommand{\arraystretch}{1.2}
    \resizebox{0.8\textwidth}{!}{\begin{tabular}{|c|l|cc|cc|cc|}
    \hline
    &\multirow{2.5}{*}{\textbf{Model}} &
    \multicolumn{2}{c|}{ACE05}&\multicolumn{2}{c|}{MAVEN}& \multicolumn{2}{c|}{KBP17} \\
    \cline{3-4}\cline{5-6}\cline{7-8}
      && Macro & Micro & Macro& Micro& Macro& Micro \\
      \hline
    \multirow{3.5}{*}{\textbf{Finetuing-based}}&Finetune&51.0$\pm$1.4&58.2$\pm$1.6&30.7$\pm$1.5&31.6$\pm$2.3&59.4$\pm$1.9&62.7$\pm$1.8\\
    &Finetune*&39.9$\pm$1.1&45.5$\pm$0.7&20.8$\pm$1.0&20.6$\pm$0.8&45.0$\pm$0.7&47.3$\pm$0.6\\
    &Pretrain+Finetune&22.9$\pm$6.0&20.3$\pm$4.3&20.9$\pm$4.6&16.9$\pm$5.2&35.1$\pm$5.9&30.1$\pm$5.5\\
    &Pretrain+Finetune*&14.6$\pm$3.3&15.6$\pm$3.4&12.5$\pm$3.8&14.9$\pm$4.0&23.4$\pm$6.8&25.8$\pm$6.3\\
    \hline
    \multirow{3.5}{*}{\textbf{Prototypical Net}}&FS-Base&63.8$\pm$2.8&67.3$\pm$2.7&44.7$\pm$1.4&44.5$\pm$2.0&65.5$\pm$2.7&67.3$\pm$3.1\\
    &FS-LexFree&52.7$\pm$2.9&53.9$\pm$3.2&25.6$\pm$1.0&21.8$\pm$1.4&60.7$\pm$2.5&61.4$\pm$2.8\\
    &FS-ClusterLoss&64.9$\pm$1.5&69.4$\pm$2.0&44.2$\pm$1.2&44.0$\pm$1.2&65.5$\pm$2.3&67.1$\pm$2.4\\
    &FS-Causal (Ours)&\textbf{73.0}$\pm$2.2&\textbf{76.9}$\pm$1.4&\textbf{52.1}$\pm$0.2&\textbf{55.0}$\pm$0.4&\textbf{70.9}$\pm$0.6&\textbf{73.2}$\pm$0.9\\
    \hline
    \multirow{3.5}{*}{\textbf{Relation Net}}&FS-Base&65.7$\pm$3.7&68.7$\pm$4.5&52.4$\pm$1.4&56.0$\pm$1.4&\textbf{67.2}$\pm$1.5&71.2$\pm$1.4\\
    &FS-LexFree &59.3$\pm$3.5&60.1$\pm$3.9&43.8$\pm$1.9&45.9$\pm$2.4&61.9$\pm$2.4&65.4$\pm$2.8\\
    &FS-ClusterLoss&57.6$\pm$2.3&60.2$\pm$3.2&46.3$\pm$1.1&51.8$\pm$1.4&56.8$\pm$3.0&62.1$\pm$2.5\\
    &FS-Causal (Ours)&\textbf{67.2}$\pm$1.4&\textbf{71.8}$\pm$1.9&\textbf{53.0}$\pm$0.5&\textbf{57.0}$\pm$0.9&66.4$\pm$0.4&\textbf{72.0}$\pm$0.6\\
    \hline
    \end{tabular}}
    \end{center}
    \caption{F1 score of 5-shot FSED on test set. * means fixing the parameters of encoder when finetuning. $\pm$ is the standard deviation of 5 random training rounds. }
    \label{tab:result}
  \end{table*}
  
  \section{Experiments}
  \subsection{Experimental Settings}
  \noindent\textbf{Datasets.\footnote{Our source codes are openly available at \href{https://github.com/chen700564/causalFSED}{https://github.com/chen700564/causalFSED}}}\quad We conducted experiments on ACE05, MAVEN \citep{wang-etal-2020-maven} and KBP17 datasets. We split train/dev/test sets according to event types and we use event types with more instances for training, the other for dev/test. To conduct 5-shot experiments, we filter event types less than 6 instances. Finally, for ACE05, its train/dev/test set contains 3598/140/149 instances and 20/10/10 types respectively, for MAVEN, those are 34651/1494/1505 instances and 120/45/45 types, for KBP17, those are 15785/768/792 instances and 25/13/13 types.
  
  \noindent\textbf{Task Settings.} Different from episode evaluation in \citet{lai-etal-2020-extensively} and \citet{cong-etal-2021-shot}, we employ a more practical event detection setting inspired by \citet{yang-katiyar-2020-simple} in Few-shot NER. We randomly sample few instances as support set and all other instances in the test set are used as queries. A support set corresponds to an event type and all types will be evaluated by traversing each event type. Models need to detection the span and type of triggers in a sentence. We also compared the results across settings in Section~\ref{sec:effect}. We evaluate all methods using macro-F1 and micro-F1 scores, and micro-F1 is taken as the primary measure.
  
  \noindent\textbf{Baselines.}\quad We conduct experiments on two metric-based methods: Prototypical Network \citep{snell2017prototypical} and Relation Network \citep{sung2018learning}, which are referred as \textbf{FS-Base}. Based on these models, we compare our causal intervention method (\textbf{FS-Casual}) with 1) \textbf{FS-LexFree} \cite{lu-etal-2019-distilling}, which address overfit triggers via adversarial learning, we use their lexical-free encoder; 2) \textbf{FS-ClusterLoss} \citep{lai-etal-2020-extensively}, which add two auxiliary loss functions when training.
  Furthermore, we compare our method with models fine-tuned with support set (\textbf{Finetune}) and pretrained using the training set (\textbf{Pretrain}). $\text{BERT}_{\text{base}}$ (uncased) is used as the encoder for all models and MLM for trigger collection.
  
  \subsection{Experimental Results}
  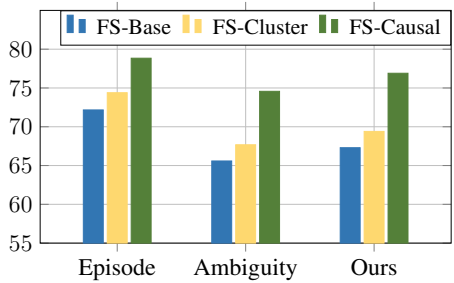
\begin{figure}
    \setlength{\abovecaptionskip}{0.2cm}
    \setlength{\belowcaptionskip}{-0.7cm}
    \begin{center}
        \resizebox{0.37\textwidth}{!}{\begin{tikzpicture}
          \begin{axis}[
            ybar,
            ymin=55,
            ymax=85,
            height=5.7 cm,
            width=8.8 cm,
            xtick=data,
            bar width=0.35cm,
            ytick={55,60,65,70,75,80},
            tick label style={font=\large}, 
            symbolic x coords={Episode, Ambiguity, Ours},
            enlarge x limits=0.3,
            legend columns=-1,
            ymajorgrids=true,
            xmajorgrids=true,
            legend style={
            at={(1,1)},
            anchor=north east,column sep=0.13cm}],
        ]
            \addplot[color={rgb,255:red,50; green,118; blue,179},fill] coordinates {(Episode, 72.17) (Ambiguity, 65.59) (Ours, 67.3)};
            \addplot[color= {rgb,255:red,254; green,216; blue,111},fill] coordinates {(Episode, 74.39) (Ambiguity, 67.68) (Ours, 69.4)};
            \addplot[color={rgb,255:red,84; green,128; blue,57},fill] coordinates {(Episode, 78.84) (Ambiguity, 74.57) (Ours, 76.9)};
    
            \legend{
                FS-Base,
                FS-Cluster,
                FS-Causal,
            }
        \end{axis}
          \end{tikzpicture}}
      \caption{Micro F1 of prototypical network with different settings on ACE05 test set.}
      \label{fig:setting}
    \end{center}
  \end{figure}

  The performance of our method and all baselines is shown in Table \ref{tab:result}. We can see that:
  
  1) \textbf{By intervening on the context in SCM and using backdoor adjustment during training, our method can effectively learn FSED models.} Compared with the original metric-based models, our method achieves 8.7\% and 1.6\% micro-F1 (average) improvement in prototypical network and relation network respectively.
  
  2) \textbf{The causal theory is a promising technique for resolving the trigger cruse problem.} Notice that FS-LexFree cannot achieve the competitive performance with the original FS models, which indicates that trigger information is import and underfitting triggers will hurt the detection performance. This verifies that trigger curse is very challenging and causal intervention can effectively resolve it.
  
  3) \textbf{Our method can achieve state-of-the-art FSED performance.} Compared with best score in baselines, our method gains 7.5\%, 1.0\%, and 2.0\% micro-F1 improvements on ACE05, MAVEN and KBP17 datasets respectively.
  
  \subsection{Effect on Different Settings}\label{sec:effect}
  To further demonstrate the effectiveness of the proposed method, we also conduct experiments under different FSED settings: 1) The primal episode-based settings (\textbf{Episode}), which is the 5+1-way 5-shot settings in \citet{lai-etal-2020-extensively}. 2) Episode + ambiguous instances (\textbf{Ambiguity}), which samples some additional negative query instances that include words same as triggers in support set to verify whether models overfit the triggers. 
 
  The performance of different models with different settings is shown in Figure \ref{fig:setting}. We can see that: 1) Generally speaking, all models can achieve better performance on \textbf{Episode} because correctly recognize high-frequent triggers can achieve good performance in this setting. Consequently, the performance under this setting can not well represent how FSED is influenced by trigger overfitting. 2) The performance of all models dropped on \textbf{Ambiguity} setting, which suggests that trigger overfitting has a significant impact on FSED. 3) Our method still maintains good performance on \textbf{Ambiguity}, which indicates that our method can alleviate the trigger curse problem by optimizing towards the underlying causality.

  \begin{table*}
    \centering
    \resizebox{0.98\textwidth}{!}{\begin{tabular}{l|l}
    \hline
    \emph{Support set 1}&I mean , I 'd like to  - -- I 'd like to see the Greens \emph{\textcolor{blue}{[Nominate]}run\textcolor{blue}{[/Nominate]}} David Cobb again.\\
    \emph{Query 1}&Release a known terrorist to run the PLO and that will bring about peace\\
    \hline
    FS-Base & Release a known terrorist to \emph{\textcolor{red}{[Nominate]}run\textcolor{red}{[/Nominate]}} the PLO and that will bring about peace\\
    FS-Causal & Release a known terrorist to run the PLO and that will bring about peace\\
    \hline
    \hline
    \emph{Support set 2}&They were \emph{\textcolor{blue}{[Suspicion]}suspected\textcolor{blue}{[/Suspicion]}} of having facilitated the suicide bomber.\\
    \emph{Query 2}&A fourth suspect, Osman Hussein, was arrested in Rome, Italy, and later extradited to the UK.\\
    \hline
    FS-Base & A fourth \emph{\textcolor{red}{[Suspicion]}suspect\textcolor{red}{[/Suspicion]}}, Osman Hussein, was arrested in Rome, Italy, and later extradited to the UK.\\
    FS-Causal & A fourth suspect, Osman Hussein, was arrested in Rome, Italy, and later extradited to the UK.\\
    \hline
    \end{tabular}}
    \caption{Ambiguous cases from ACE05 and MAVEN test set. The results are based on prototypical network and \emph{Support set} means one instance in the support set.}
    \label{tab:case}
  \end{table*}

  \subsection{Case Study}
  We select ambiguous cases (in Table \ref{tab:case}) to better illustrate the effectiveness of our method. For \emph{Query 1}, FS-Base wrongly detects the word \emph{run} to be a trigger word. In \emph{Support set 1}, \emph{run} means nominating while \emph{run} means managing in \emph{Query 1}. FS-Base fails to recognize such different sense of word under context. For \emph{Query 2}, FS-Base makes mistake again on the ambiguous word \emph{suspect}. Even though \emph{suspect} is the noun form of \emph{suspected} in \emph{Support set 2}, it does not trigger a \texttt{Suspicion} event in \emph{Query 2}. In contract to FS-Base, our approach is able to handle both cases correctly, illustrating its effectiveness.

  \section{Related Work}
  \noindent\textbf{Causal Inference.}  Causal inference aims to make reliable predictions using the causal effect between variables \citep{pearl2009causality}. Many studies have used causal theory to improve model robustness \citep{wang2020visual,wang2020visual2,qi2020two,tang2020unbiased,zeng-etal-2020-counterfactual}. Recently, backdoor adjustment has been used to remove the spurious association brought by the confounder \citep{tang2020long,zhang2020causal,yue2020interventional,liu-etal-2021-element,zhang-etal-2021-de}. 
  
  \noindent\textbf{Few-shot Event Detection.} Few-shot event detection has been studied in many different settings. \citet{bronstein-etal-2015-seed} collect some seed triggers, then detect unseen event with feature-based method. \citet{deng2020meta} decompose FSED into two sub-tasks: trigger identification and few-shot classification. \citet{feng2020probing} adopt a sentence-level few-shot classification without triggers. \citet{lai-etal-2020-extensively} and \citet{cong-etal-2021-shot} adopt N+1-way few-shot setting that is closest to our setting.
  
  \section{Conclusions}
  This paper proposes to revisit the trigger curse in FSED from a causal view. Specifically, we identify the cause of the trigger curse problem from a structural causal model, and then solve the problem through casual intervention via backdoor adjustment. Experimental results demonstrate the effectiveness and robustness of our methods.

  \section*{Acknowledgments} 
  
  We thank the reviewers for their insightful comments and helpful suggestions. This research work is supported by National Key R\&D Program of China under Grant 2018YFB1005100, the National Natural Science Foundation of China under Grants no.  62106251 and 62076233, and in part by the Youth Innovation Promotion Association CAS(2018141).
  
\bibliography{custom}

\begin{thebibliography}{49}
\expandafter\ifx\csname natexlab\endcsname\relax\def\natexlab#1{#1}\fi

\bibitem[{Banarescu et~al.(2013)Banarescu, Bonial, Cai, Georgescu, Griffitt,
  Hermjakob, Knight, Koehn, Palmer, and
  Schneider}]{banarescu-etal-2013-abstract}
Laura Banarescu, Claire Bonial, Shu Cai, Madalina Georgescu, Kira Griffitt, Ulf
  Hermjakob, Kevin Knight, Philipp Koehn, Martha Palmer, and Nathan Schneider.
  2013.
\newblock \href {https://www.aclweb.org/anthology/W13-2322} {{A}bstract
  {M}eaning {R}epresentation for sembanking}.
\newblock In \emph{Proceedings of the 7th Linguistic Annotation Workshop and
  Interoperability with Discourse}, pages 178--186, Sofia, Bulgaria.
  Association for Computational Linguistics.

\bibitem[{Bronstein et~al.(2015)Bronstein, Dagan, Li, Ji, and
  Frank}]{bronstein-etal-2015-seed}
Ofer Bronstein, Ido Dagan, Qi~Li, Heng Ji, and Anette Frank. 2015.
\newblock \href {https://doi.org/10.3115/v1/P15-2061} {Seed-based event trigger
  labeling: How far can event descriptions get us?}
\newblock In \emph{Proceedings of the 53rd Annual Meeting of the Association
  for Computational Linguistics and the 7th International Joint Conference on
  Natural Language Processing (Volume 2: Short Papers)}, pages 372--376,
  Beijing, China. Association for Computational Linguistics.

\bibitem[{Chen et~al.(2015)Chen, Xu, Liu, Zeng, and
  Zhao}]{chen-etal-2015-event}
Yubo Chen, Liheng Xu, Kang Liu, Daojian Zeng, and Jun Zhao. 2015.
\newblock \href {https://doi.org/10.3115/v1/P15-1017} {Event extraction via
  dynamic multi-pooling convolutional neural networks}.
\newblock In \emph{Proceedings of the 53rd Annual Meeting of the Association
  for Computational Linguistics and the 7th International Joint Conference on
  Natural Language Processing (Volume 1: Long Papers)}, pages 167--176,
  Beijing, China. Association for Computational Linguistics.

\bibitem[{Chen et~al.(2018)Chen, Yang, Liu, Zhao, and
  Jia}]{chen-etal-2018-collective}
Yubo Chen, Hang Yang, Kang Liu, Jun Zhao, and Yantao Jia. 2018.
\newblock \href {https://doi.org/10.18653/v1/D18-1158} {Collective event
  detection via a hierarchical and bias tagging networks with gated multi-level
  attention mechanisms}.
\newblock In \emph{Proceedings of the 2018 Conference on Empirical Methods in
  Natural Language Processing}, pages 1267--1276, Brussels, Belgium.
  Association for Computational Linguistics.

\bibitem[{Cong et~al.(2021)Cong, Cui, Yu, Liu, Yubin, and
  Wang}]{cong-etal-2021-shot}
Xin Cong, Shiyao Cui, Bowen Yu, Tingwen Liu, Wang Yubin, and Bin Wang. 2021.
\newblock \href {https://doi.org/10.18653/v1/2021.findings-acl.3} {{F}ew-{S}hot
  {E}vent {D}etection with {P}rototypical {A}mortized {C}onditional {R}andom
  {F}ield}.
\newblock In \emph{Findings of the Association for Computational Linguistics:
  ACL-IJCNLP 2021}, pages 28--40, Online. Association for Computational
  Linguistics.

\bibitem[{Deng et~al.(2020)Deng, Zhang, Kang, Zhang, Zhang, and
  Chen}]{deng2020meta}
Shumin Deng, Ningyu Zhang, Jiaojian Kang, Yichi Zhang, Wei Zhang, and Huajun
  Chen. 2020.
\newblock Meta-learning with dynamic-memory-based prototypical network for
  few-shot event detection.
\newblock In \emph{Proceedings of the 13th International Conference on Web
  Search and Data Mining}, pages 151--159.

\bibitem[{Devlin et~al.(2019)Devlin, Chang, Lee, and
  Toutanova}]{devlin-etal-2019-bert}
Jacob Devlin, Ming-Wei Chang, Kenton Lee, and Kristina Toutanova. 2019.
\newblock \href {https://doi.org/10.18653/v1/N19-1423} {{BERT}: Pre-training of
  deep bidirectional transformers for language understanding}.
\newblock In \emph{Proceedings of the 2019 Conference of the North {A}merican
  Chapter of the Association for Computational Linguistics: Human Language
  Technologies, Volume 1 (Long and Short Papers)}, pages 4171--4186,
  Minneapolis, Minnesota. Association for Computational Linguistics.

\bibitem[{Doddington et~al.(2004)Doddington, Mitchell, Przybocki, Ramshaw,
  Strassel, and Weischedel}]{doddington2004automatic}
George~R Doddington, Alexis Mitchell, Mark~A Przybocki, Lance~A Ramshaw,
  Stephanie~M Strassel, and Ralph~M Weischedel. 2004.
\newblock The automatic content extraction (ace) program-tasks, data, and
  evaluation.
\newblock In \emph{Lrec}, volume~2, pages 837--840. Lisbon.

\bibitem[{Du and Cardie(2020)}]{du-cardie-2020-event}
Xinya Du and Claire Cardie. 2020.
\newblock \href {https://doi.org/10.18653/v1/2020.emnlp-main.49} {Event
  extraction by answering (almost) natural questions}.
\newblock In \emph{Proceedings of the 2020 Conference on Empirical Methods in
  Natural Language Processing (EMNLP)}, pages 671--683, Online. Association for
  Computational Linguistics.

\bibitem[{Duan et~al.(2017)Duan, He, and Zhao}]{duan2017exploiting}
Shaoyang Duan, Ruifang He, and Wenli Zhao. 2017.
\newblock Exploiting document level information to improve event detection via
  recurrent neural networks.
\newblock In \emph{Proceedings of the Eighth International Joint Conference on
  Natural Language Processing (Volume 1: Long Papers)}, pages 352--361.

\bibitem[{Feng et~al.(2020)Feng, Yuan, and Zhang}]{feng2020probing}
Rui Feng, Jie Yuan, and Chao Zhang. 2020.
\newblock \href {http://arxiv.org/abs/2010.11325} {Probing and fine-tuning
  reading comprehension models for few-shot event extraction}.
\newblock \emph{CoRR}, abs/2010.11325.

\bibitem[{Fillmore(1967)}]{fillmore1967case}
Charles Fillmore. 1967.
\newblock The case for case.

\bibitem[{Gardner et~al.(2018)Gardner, Grus, Neumann, Tafjord, Dasigi, Liu,
  Peters, Schmitz, and Zettlemoyer}]{gardner2018allennlp}
Matt Gardner, Joel Grus, Mark Neumann, Oyvind Tafjord, Pradeep Dasigi,
  Nelson~F. Liu, Matthew~E. Peters, Michael Schmitz, and Luke Zettlemoyer.
  2018.
\newblock \href {http://arxiv.org/abs/1803.07640} {Allennlp: {A} deep semantic
  natural language processing platform}.
\newblock \emph{CoRR}, abs/1803.07640.

\bibitem[{Gildea and Jurafsky(2002)}]{gildea2002automatic}
Daniel Gildea and Daniel Jurafsky. 2002.
\newblock Automatic labeling of semantic roles.
\newblock \emph{Computational linguistics}, 28(3):245--288.

\bibitem[{Ji and Grishman(2008)}]{ji-grishman-2008-refining}
Heng Ji and Ralph Grishman. 2008.
\newblock \href {https://www.aclweb.org/anthology/P08-1030} {Refining event
  extraction through cross-document inference}.
\newblock In \emph{Proceedings of ACL-08: HLT}, pages 254--262, Columbus, Ohio.
  Association for Computational Linguistics.

\bibitem[{Ji et~al.(2019)Ji, Lin, Gao, and Wan}]{ji-etal-2019-exploiting}
Yuze Ji, Youfang Lin, Jianwei Gao, and Huaiyu Wan. 2019.
\newblock \href {https://doi.org/10.18653/v1/K19-1057} {Exploiting the entity
  type sequence to benefit event detection}.
\newblock In \emph{Proceedings of the 23rd Conference on Computational Natural
  Language Learning (CoNLL)}, pages 613--623, Hong Kong, China. Association for
  Computational Linguistics.

\bibitem[{Lai et~al.(2020)Lai, Nguyen, and
  Dernoncourt}]{lai-etal-2020-extensively}
Viet~Dac Lai, Thien~Huu Nguyen, and Franck Dernoncourt. 2020.
\newblock \href {https://doi.org/10.18653/v1/2020.nuse-1.5} {Extensively
  matching for few-shot learning event detection}.
\newblock In \emph{Proceedings of the First Joint Workshop on Narrative
  Understanding, Storylines, and Events}, pages 38--45, Online. Association for
  Computational Linguistics.

\bibitem[{Liao and Grishman(2010)}]{liao-grishman-2010-using}
Shasha Liao and Ralph Grishman. 2010.
\newblock \href {https://www.aclweb.org/anthology/P10-1081} {Using document
  level cross-event inference to improve event extraction}.
\newblock In \emph{Proceedings of the 48th Annual Meeting of the Association
  for Computational Linguistics}, pages 789--797, Uppsala, Sweden. Association
  for Computational Linguistics.

\bibitem[{Lin et~al.(2018)Lin, Lu, Han, and Sun}]{lin-etal-2018-nugget}
Hongyu Lin, Yaojie Lu, Xianpei Han, and Le~Sun. 2018.
\newblock \href {https://doi.org/10.18653/v1/P18-1145} {Nugget proposal
  networks for {C}hinese event detection}.
\newblock In \emph{Proceedings of the 56th Annual Meeting of the Association
  for Computational Linguistics (Volume 1: Long Papers)}, pages 1565--1574,
  Melbourne, Australia. Association for Computational Linguistics.

\bibitem[{Lin et~al.(2019{\natexlab{a}})Lin, Lu, Han, and
  Sun}]{lin-etal-2019-cost}
Hongyu Lin, Yaojie Lu, Xianpei Han, and Le~Sun. 2019{\natexlab{a}}.
\newblock \href {https://doi.org/10.18653/v1/P19-1521} {Cost-sensitive
  regularization for label confusion-aware event detection}.
\newblock In \emph{Proceedings of the 57th Annual Meeting of the Association
  for Computational Linguistics}, pages 5278--5283, Florence, Italy.
  Association for Computational Linguistics.

\bibitem[{Lin et~al.(2019{\natexlab{b}})Lin, Lu, Han, and
  Sun}]{lin-etal-2019-sequence}
Hongyu Lin, Yaojie Lu, Xianpei Han, and Le~Sun. 2019{\natexlab{b}}.
\newblock \href {https://doi.org/10.18653/v1/P19-1511} {Sequence-to-nuggets:
  Nested entity mention detection via anchor-region networks}.
\newblock In \emph{Proceedings of the 57th Annual Meeting of the Association
  for Computational Linguistics}, pages 5182--5192, Florence, Italy.
  Association for Computational Linguistics.

\bibitem[{Liu et~al.(2021)Liu, Yan, Lin, Han, and Sun}]{liu-etal-2021-element}
Fangchao Liu, Lingyong Yan, Hongyu Lin, Xianpei Han, and Le~Sun. 2021.
\newblock \href {https://doi.org/10.18653/v1/2021.acl-long.361} {Element
  intervention for open relation extraction}.
\newblock In \emph{Proceedings of the 59th Annual Meeting of the Association
  for Computational Linguistics and the 11th International Joint Conference on
  Natural Language Processing (Volume 1: Long Papers)}, pages 4683--4693,
  Online. Association for Computational Linguistics.

\bibitem[{Liu et~al.(2019)Liu, Chen, and Liu}]{liu2019exploiting}
Jian Liu, Yubo Chen, and Kang Liu. 2019.
\newblock Exploiting the ground-truth: An adversarial imitation based knowledge
  distillation approach for event detection.
\newblock In \emph{Proceedings of the AAAI Conference on Artificial
  Intelligence}, volume~33, pages 6754--6761.

\bibitem[{Liu et~al.(2020{\natexlab{a}})Liu, Chen, Liu, Bi, and
  Liu}]{liu-etal-2020-mrc}
Jian Liu, Yubo Chen, Kang Liu, Wei Bi, and Xiaojiang Liu. 2020{\natexlab{a}}.
\newblock \href {https://doi.org/10.18653/v1/2020.emnlp-main.128} {Event
  extraction as machine reading comprehension}.
\newblock In \emph{Proceedings of the 2020 Conference on Empirical Methods in
  Natural Language Processing (EMNLP)}, pages 1641--1651, Online. Association
  for Computational Linguistics.

\bibitem[{Liu et~al.(2020{\natexlab{b}})Liu, Chen, Liu, Jia, and
  Sheng}]{liu-etal-2020-context}
Jian Liu, Yubo Chen, Kang Liu, Yantao Jia, and Zhicheng Sheng.
  2020{\natexlab{b}}.
\newblock \href {https://www.aclweb.org/anthology/2020.findings-emnlp.229} {How
  does context matter? on the robustness of event detection with
  context-selective mask generalization}.
\newblock In \emph{Findings of the Association for Computational Linguistics:
  EMNLP 2020}, pages 2523--2532, Online. Association for Computational
  Linguistics.

\bibitem[{Liu et~al.(2017)Liu, Chen, Liu, and Zhao}]{liu-etal-2017-exploiting}
Shulin Liu, Yubo Chen, Kang Liu, and Jun Zhao. 2017.
\newblock \href {https://doi.org/10.18653/v1/P17-1164} {Exploiting argument
  information to improve event detection via supervised attention mechanisms}.
\newblock In \emph{Proceedings of the 55th Annual Meeting of the Association
  for Computational Linguistics (Volume 1: Long Papers)}, pages 1789--1798,
  Vancouver, Canada. Association for Computational Linguistics.

\bibitem[{Lu et~al.(2019)Lu, Lin, Han, and Sun}]{lu-etal-2019-distilling}
Yaojie Lu, Hongyu Lin, Xianpei Han, and Le~Sun. 2019.
\newblock \href {https://doi.org/10.18653/v1/P19-1429} {Distilling
  discrimination and generalization knowledge for event detection via
  delta-representation learning}.
\newblock In \emph{Proceedings of the 57th Annual Meeting of the Association
  for Computational Linguistics}, pages 4366--4376, Florence, Italy.
  Association for Computational Linguistics.

\bibitem[{Lu et~al.(2021)Lu, Lin, Xu, Han, Tang, Li, Sun, Liao, and
  Chen}]{lu-etal-2021-text2event}
Yaojie Lu, Hongyu Lin, Jin Xu, Xianpei Han, Jialong Tang, Annan Li, Le~Sun,
  Meng Liao, and Shaoyi Chen. 2021.
\newblock \href {https://doi.org/10.18653/v1/2021.acl-long.217}
  {{T}ext2{E}vent: Controllable sequence-to-structure generation for end-to-end
  event extraction}.
\newblock In \emph{Proceedings of the 59th Annual Meeting of the Association
  for Computational Linguistics and the 11th International Joint Conference on
  Natural Language Processing (Volume 1: Long Papers)}, pages 2795--2806,
  Online. Association for Computational Linguistics.

\bibitem[{Nguyen et~al.(2016)Nguyen, Cho, and
  Grishman}]{nguyen-etal-2016-joint}
Thien~Huu Nguyen, Kyunghyun Cho, and Ralph Grishman. 2016.
\newblock \href {https://doi.org/10.18653/v1/N16-1034} {Joint event extraction
  via recurrent neural networks}.
\newblock In \emph{Proceedings of the 2016 Conference of the North {A}merican
  Chapter of the Association for Computational Linguistics: Human Language
  Technologies}, pages 300--309, San Diego, California. Association for
  Computational Linguistics.

\bibitem[{Nguyen and Grishman(2015)}]{nguyen-grishman-2015-event}
Thien~Huu Nguyen and Ralph Grishman. 2015.
\newblock \href {https://doi.org/10.3115/v1/P15-2060} {Event detection and
  domain adaptation with convolutional neural networks}.
\newblock In \emph{Proceedings of the 53rd Annual Meeting of the Association
  for Computational Linguistics and the 7th International Joint Conference on
  Natural Language Processing (Volume 2: Short Papers)}, pages 365--371,
  Beijing, China. Association for Computational Linguistics.

\bibitem[{Pearl(1995)}]{pearl1995causal}
Judea Pearl. 1995.
\newblock Causal diagrams for empirical research.
\newblock \emph{Biometrika}, 82(4):669--688.

\bibitem[{Pearl(2009)}]{pearl2009causality}
Judea Pearl. 2009.
\newblock \emph{Causality}.
\newblock Cambridge university press.

\bibitem[{Pearl(2014)}]{pearl2014probabilistic}
Judea Pearl. 2014.
\newblock \emph{Probabilistic reasoning in intelligent systems: networks of
  plausible inference}.
\newblock Elsevier.

\bibitem[{Pearl et~al.(2016)Pearl, Glymour, and Jewell}]{pearl2016causal}
Judea Pearl, Madelyn Glymour, and Nicholas~P Jewell. 2016.
\newblock \emph{Causal inference in statistics: A primer}.
\newblock John Wiley \& Sons.

\bibitem[{Qi et~al.(2020)Qi, Niu, Huang, and Zhang}]{qi2020two}
Jiaxin Qi, Yulei Niu, Jianqiang Huang, and Hanwang Zhang. 2020.
\newblock Two causal principles for improving visual dialog.
\newblock In \emph{Proceedings of the IEEE/CVF Conference on Computer Vision
  and Pattern Recognition}, pages 10860--10869.

\bibitem[{Snell et~al.(2017)Snell, Swersky, and Zemel}]{snell2017prototypical}
Jake Snell, Kevin Swersky, and Richard Zemel. 2017.
\newblock Prototypical networks for few-shot learning.
\newblock In \emph{Advances in neural information processing systems}, pages
  4077--4087.

\bibitem[{Sung et~al.(2018)Sung, Yang, Zhang, Xiang, Torr, and
  Hospedales}]{sung2018learning}
Flood Sung, Yongxin Yang, Li~Zhang, Tao Xiang, Philip~HS Torr, and Timothy~M
  Hospedales. 2018.
\newblock Learning to compare: Relation network for few-shot learning.
\newblock In \emph{Proceedings of the IEEE Conference on Computer Vision and
  Pattern Recognition}, pages 1199--1208.

\bibitem[{Tang et~al.(2020{\natexlab{a}})Tang, Huang, and Zhang}]{tang2020long}
Kaihua Tang, Jianqiang Huang, and Hanwang Zhang. 2020{\natexlab{a}}.
\newblock \href
  {https://proceedings.neurips.cc/paper/2020/hash/1091660f3dff84fd648efe31391c5524-Abstract.html}
  {Long-tailed classification by keeping the good and removing the bad momentum
  causal effect}.
\newblock In \emph{Advances in Neural Information Processing Systems 33: Annual
  Conference on Neural Information Processing Systems 2020, NeurIPS 2020,
  December 6-12, 2020, virtual}.

\bibitem[{Tang et~al.(2020{\natexlab{b}})Tang, Niu, Huang, Shi, and
  Zhang}]{tang2020unbiased}
Kaihua Tang, Yulei Niu, Jianqiang Huang, Jiaxin Shi, and Hanwang Zhang.
  2020{\natexlab{b}}.
\newblock Unbiased scene graph generation from biased training.
\newblock In \emph{Proceedings of the IEEE/CVF Conference on Computer Vision
  and Pattern Recognition}, pages 3716--3725.

\bibitem[{Taylor(1953)}]{taylor1953cloze}
Wilson~L Taylor. 1953.
\newblock “cloze procedure”: A new tool for measuring readability.
\newblock \emph{Journalism quarterly}, 30(4):415--433.

\bibitem[{Wang et~al.(2020{\natexlab{a}})Wang, Huang, Zhang, and
  Sun}]{wang2020visual}
Tan Wang, Jianqiang Huang, Hanwang Zhang, and Qianru Sun. 2020{\natexlab{a}}.
\newblock Visual commonsense r-cnn.
\newblock In \emph{Proceedings of the IEEE/CVF Conference on Computer Vision
  and Pattern Recognition}, pages 10760--10770.

\bibitem[{Wang et~al.(2020{\natexlab{b}})Wang, Huang, Zhang, and
  Sun}]{wang2020visual2}
Tan Wang, Jianqiang Huang, Hanwang Zhang, and Qianru Sun. 2020{\natexlab{b}}.
\newblock Visual commonsense representation learning via causal inference.
\newblock In \emph{Proceedings of the IEEE/CVF Conference on Computer Vision
  and Pattern Recognition Workshops}, pages 378--379.

\bibitem[{Wang et~al.(2020{\natexlab{c}})Wang, Wang, Han, Jiang, Han, Liu, Li,
  Li, Lin, and Zhou}]{wang-etal-2020-maven}
Xiaozhi Wang, Ziqi Wang, Xu~Han, Wangyi Jiang, Rong Han, Zhiyuan Liu, Juanzi
  Li, Peng Li, Yankai Lin, and Jie Zhou. 2020{\natexlab{c}}.
\newblock \href {https://www.aclweb.org/anthology/2020.emnlp-main.129}
  {{MAVEN}: {A} {M}assive {G}eneral {D}omain {E}vent {D}etection {D}ataset}.
\newblock In \emph{Proceedings of the 2020 Conference on Empirical Methods in
  Natural Language Processing (EMNLP)}, pages 1652--1671, Online. Association
  for Computational Linguistics.

\bibitem[{Wolf et~al.(2019)Wolf, Debut, Sanh, Chaumond, Delangue, Moi, Cistac,
  Rault, Louf, Funtowicz, and Brew}]{wolf2019huggingface}
Thomas Wolf, Lysandre Debut, Victor Sanh, Julien Chaumond, Clement Delangue,
  Anthony Moi, Pierric Cistac, Tim Rault, R{\'{e}}mi Louf, Morgan Funtowicz,
  and Jamie Brew. 2019.
\newblock \href {http://arxiv.org/abs/1910.03771} {Huggingface's transformers:
  State-of-the-art natural language processing}.
\newblock \emph{CoRR}, abs/1910.03771.

\bibitem[{Yang and Katiyar(2020)}]{yang-katiyar-2020-simple}
Yi~Yang and Arzoo Katiyar. 2020.
\newblock \href {https://www.aclweb.org/anthology/2020.emnlp-main.516} {Simple
  and effective few-shot named entity recognition with structured nearest
  neighbor learning}.
\newblock In \emph{Proceedings of the 2020 Conference on Empirical Methods in
  Natural Language Processing (EMNLP)}, pages 6365--6375, Online. Association
  for Computational Linguistics.

\bibitem[{Yue et~al.(2020)Yue, Zhang, Sun, and Hua}]{yue2020interventional}
Zhongqi Yue, Hanwang Zhang, Qianru Sun, and Xian{-}Sheng Hua. 2020.
\newblock \href
  {https://proceedings.neurips.cc/paper/2020/hash/1cc8a8ea51cd0adddf5dab504a285915-Abstract.html}
  {Interventional few-shot learning}.
\newblock In \emph{Advances in Neural Information Processing Systems 33: Annual
  Conference on Neural Information Processing Systems 2020, NeurIPS 2020,
  December 6-12, 2020, virtual}.

\bibitem[{Zeng et~al.(2020)Zeng, Li, Zhai, and
  Zhang}]{zeng-etal-2020-counterfactual}
Xiangji Zeng, Yunliang Li, Yuchen Zhai, and Yin Zhang. 2020.
\newblock \href {https://doi.org/10.18653/v1/2020.emnlp-main.590}
  {Counterfactual generator: A weakly-supervised method for named entity
  recognition}.
\newblock In \emph{Proceedings of the 2020 Conference on Empirical Methods in
  Natural Language Processing (EMNLP)}, pages 7270--7280, Online. Association
  for Computational Linguistics.

\bibitem[{Zhang et~al.(2020)Zhang, Zhang, Tang, Hua, and Sun}]{zhang2020causal}
Dong Zhang, Hanwang Zhang, Jinhui Tang, Xian{-}Sheng Hua, and Qianru Sun. 2020.
\newblock \href
  {https://proceedings.neurips.cc/paper/2020/hash/07211688a0869d995947a8fb11b215d6-Abstract.html}
  {Causal intervention for weakly-supervised semantic segmentation}.
\newblock In \emph{Advances in Neural Information Processing Systems 33: Annual
  Conference on Neural Information Processing Systems 2020, NeurIPS 2020,
  December 6-12, 2020, virtual}.

\bibitem[{Zhang et~al.(2021)Zhang, Lin, Han, and Sun}]{zhang-etal-2021-de}
Wenkai Zhang, Hongyu Lin, Xianpei Han, and Le~Sun. 2021.
\newblock \href {https://doi.org/10.18653/v1/2021.acl-long.371} {De-biasing
  distantly supervised named entity recognition via causal intervention}.
\newblock In \emph{Proceedings of the 59th Annual Meeting of the Association
  for Computational Linguistics and the 11th International Joint Conference on
  Natural Language Processing (Volume 1: Long Papers)}, pages 4803--4813,
  Online. Association for Computational Linguistics.

\end{thebibliography}
\bibliographystyle{acl_natbib}
\newpage
\appendix
\newcommand\independent{\protect\mathpalette{\protect\independenT}{\perp}}
\def\independenT#1#2{\mathrel{\rlap{$#1#2$}\mkern2mu{#1#2}}}

\section{Proof of Backdoor Adjustment}
  We prove the backdoor adjustment for our SCM using the rules of do-calculus \citep{pearl1995causal}.
  
  For a causal graph $G$, let $G_{\overline{X}}$ denote the graph where all of the incoming edges to Node $X$ are removed. let $G_{\underline{X}}$ denote the graph where all of the outgoing edges from Node $X$ are removed. $\independent_G$ denotes d-separation in $G$.
  
  \textbf{D-separation} \citep{pearl2014probabilistic}: Two (sets of) nodes $X$ and $Y$ are d-separation by a set of nodes $Z$ (i.e. $X \independent_G Y | Z$) if all of the paths between (any node in) $X$ and (any node in) $Y$ are blocked by $Z$. 
  
  The rules of do-calculus are:
  
  \textbf{Rule 1}
  \begin{equation*}
    \begin{aligned}
      P(y|do(t),z,w)&=P(y|do(t),w)\\
    &\text{if}\quad Y\independent_{G_{\overline{T}}} Z|T,W
    \end{aligned}
  \end{equation*}
  
  \textbf{Rule 2}
  \begin{equation*}
    \begin{aligned}
      P(y|do(t),do(z),w)&=P(y|do(t),z,w)\\
    &\text{if}\quad Y\independent_{G_{\overline{T}\underline{Z}}} Z|T,W
    \end{aligned}
  \end{equation*}
  
  \textbf{Rule 3}
  \begin{equation}
    \begin{aligned}
      P(y|do(t),do(z),w)&=P(y|do(t),w)\\
    &\text{if}\quad Y\independent_{G_{\overline{T}\underline{Z(W)}}} Z|T,W
    \end{aligned}
  \end{equation}
  where $Z(W)$ denotes the set of nodes of Z that aren't ancestors of any node of $W$ in $G_{\overline{T}}$.
  
  We can prove our interventional distribution $P(Y|do(C=\mathcal{C}),E=e)$:
  
  \textbf{Step 1} Using the law of total probability:
  $$
  \begin{aligned}
    &P(Y|do(C=\mathcal{C}),E=e,Q=q)\\
    =&\sum\limits_{t\in T}\sum\limits_{s\in S} [P(Y|do(\mathcal{C}),e,t,s,q) \times\\
    &P(s,t|do(\mathcal{C}),e,q)]\\
  \end{aligned}
  $$
  
  \textbf{Step 2} Using the law of conditional probability:
  $$
  \begin{aligned}
    &P(Y|do(C=\mathcal{C}),E=e,Q=q)\\
    =&\sum\limits_{t\in T}\sum\limits_{s\in S}[ P(Y|do(\mathcal{C}),e,t,s,q) \times\\
    & P(s|do(\mathcal{C}),e,t,q)P(t|do(\mathcal{C}),e,q)]\\
  \end{aligned}
  $$
  
  \textbf{Step 3} Using the \textbf{Rule 3}:
  $$
  \begin{aligned}
    &P(Y|do(C=\mathcal{C}),E=e,Q=q)\\
    =&\sum\limits_{t\in T}\sum\limits_{s\in S} [P(Y|e,t,s,q)\times\\
    &P(s|do(\mathcal{C}),e,t,q)P(t|e,q)]\\
  \end{aligned}
  $$
  
  \textbf{Step 4} Using the \textbf{Rule 1}:
  $$
  \begin{aligned}
    &P(Y|do(C=\mathcal{C}),E=e,Q=q)\\
    =&\sum\limits_{t\in T}\sum\limits_{s\in S} P(Y|s,q)P(s|do(\mathcal{C}),t)P(t|e)\\
  \end{aligned}
  $$
  
  \textbf{Step 5} Using the \textbf{Rule 2}:
  $$
  \begin{aligned}
    &P(Y|do(C=\mathcal{C}),E=e,Q=q)\\
    =&\sum\limits_{t\in T}\sum\limits_{s\in S} P(Y|s,q)P(s|\mathcal{C},t)P(t|e)\\
  \end{aligned}
  $$
  
  \section{Detailed Task Settings}
  \label{sec:setting}
  \paragraph{One-way K-Shot Settings.}We adopt One-way K-shot setting in our experiments, in which the support set in an episode contains one event type (called concerned event) and the query can contain any event type. The model aims to detect triggers of the concerned event in query and all types will be evaluated by traversing each event type. The support set and query in an episode can be formulated as follows:
  $$\mathcal{S}=\{(S^1,E,Y^1),\ldots,(S^K,E,Y^K)\}$$
  where $\mathcal{S}$ is the support set, $E$ is the concerned event, $S^i=\{s_1^i,s_2^i,\ldots,s_{n_i}^i\}$ is the i-th sentence in support, $s_j^i$ is the j-th token in $S^i$, $Y^i=\{y_1^i,y_2^i,\ldots,y_n^i\}$ is the labels of tokens in $S_i$ and $y_j^i = 1$ only if $t_i$ is the trigger (or part of trigger) of concerned event, otherwise $y_j^i=0$.
  
  $$\mathcal{Q}=\{Q^1,Q^2,\ldots,Q^M\}$$
  where $\mathcal{Q}$ is the set of query and $Q^i=\{q_1^i,q_2^i,\ldots,q_{m_i}^i\}$ is the i-th query sentence and $q_j^i$ is the j-th token in $Q^i$
  
  The model is expected to output the concerned event in $\mathcal{Q}$:
  $$\begin{aligned}
    \mathcal{O}_\mathcal{Q}=\{&(Q^1,E,T^{1}_{1}),\ldots,(Q^1,E,T^{1}_{n_1}),\\
    &(Q^2,E,T^{2}_{1}),\ldots,(Q^2,E,T^{2}_{n_2}),\\
    &\ldots,\\
    &(Q^M,E,T^{M}_{1}),\ldots,(Q^M,E,T^{M}_{n_M})\}
  \end{aligned}$$
  where $\mathcal{O}_\mathcal{Q}$ is the set of triggers of concerned event detected in $\mathcal{Q}$, $T_k^{i}$ is the k-th trigger of concerned event in sentence $Q^i$ and $n_i \ge 0$ means the number of triggers of concerned event in $Q^i$.
  
  \paragraph{Evaluation}We improve the traditional episode evaluation setting by evaluating the full test set. For each event type in test set, we randomly sample K instances as support set and all other instances are used as query. Following previous event detection works \citep{chen-etal-2015-event}, the predicted trigger is correct if its event type and offsets match those of a gold trigger. We evaluate all methods using macro-F1 and micro-F1 scores, and micro-F1 is taken as the primary measure.

  \section{Few-shot Event Detection Baselines}
  \label{sec:model-baseline}
  We use two metric-base methods in our experiments: Prototypical network \citep{snell2017prototypical} and Relation network \citep{sung2018learning}, which contain an encoder component and a classifier component.
  
  \paragraph{Encoder} We use BERT \citep{devlin-etal-2019-bert} to encoder the support set and the query. Given a sentece $\mathbf{X}=\{x_1,x_2,\ldots,x_n\}$, BERT encodes the sequence and output the represent of each token in $\mathbf{X}$: $\mathbf{R} = \{\mathbf{r_1},\mathbf{r_2},\ldots,\mathbf{r_{n}}\}$. After obtaining the feature representation of the support set, we calculate the prototype of the categories (concerned event and other):
  $$
    \mathbf{p}_i=\frac{1}{|\mathcal{R}_i|}\sum\limits_{\mathbf{r}\in \mathcal{R}_i}\mathbf{r},\quad i=0,1
  $$
  where $\mathbf{p}_i$ is the prototype of category $i$, $\mathcal{R}_i$ is the set of feature representation of tokens that labeled with $y=i$ in support set.
  
  \paragraph{Classifier} The models classify each token in query based on its similarity to the prototype.
  
  We first calculate the similarity between prototype and token in query.
  \begin{equation}
    \label{eq:sim}
  s_{i,j,k} = g(\mathbf{p}_k,\mathbf{q}^i_j),\quad k=0,1
  \end{equation}
  where $g(\textbf{x},\textbf{y})$ measures the similarity between $\textbf{x}$ and $\textbf{y}$, $\mathbf{q}^i_j$ is the represent of j-th token in i-th query sentence.
  
  Then we calculate the probability distribution of token $q^i_j$:
  \begin{equation}
    \label{eq:p}
    P(Y|q^i_j,\mathcal{S}) = \text{Softmax}(s_{i,j,0},s_{i,j,1})
  \end{equation}
  
  During training, we use the Cross-Entropy loss on each token of query. And the support set and the query are randomly sampled from the training set.
  
  When evaluating, we treat the labels as IO tagging schemes, and adjacent I are considered to be the same trigger so that we can handle a trigger with multiple tokens.
  
  \paragraph{Similarity Functions} For prototypical network, the similarity in Equation \ref{eq:sim} is Euclidean distance. For relation network, we calculate similarity using neural networks. Unlike the original paper, we find the following calculation to be more efficient: $g(\mathbf{p}_k,\mathbf{q}_i^j) = F(\mathbf{p}_k \oplus \mathbf{q}_i^j \oplus |\mathbf{p}_k-\mathbf{q}_i^j| )$ where $\oplus$ means concatenation vectors and $F$ is two-layer feed-forward neural networks with a ReLU function on the first layer.
  
  \section{Proof of Loss Function}
  We prove $\mathcal{L}_{SG}(\theta)$ is equivalent to $\mathcal{L(\theta)}$, which indicates that minimizing $\mathcal{L}_{SG}(\theta)$ is equivalent to minimizing $\mathcal{L(\theta)}$. At first , we define a function $\phi(s,q) \propto P(Y|s,q;\theta)$ and then we need to prove that $g(\sum_{t\in T}\sum_{s\in S} P(t|e) p(s|\mathcal{C},t)\mathbf{r}_s,\mathbf{q}) = f(\sum_{t\in T}\sum_{s\in S} P(t|e) P(s|\mathcal{C},t) \phi(s,q))$.
  \begin{table*}
    \begin{center}
    \renewcommand{\arraystretch}{1.2}
    \resizebox{0.65\textwidth}{!}{\begin{tabular}{|l|cc|cc|cc|}
    \hline
    \multirow{2.5}{*}{\textbf{Model}} &
    \multicolumn{2}{c|}{ACE05}&\multicolumn{2}{c|}{MAVEN}&\multicolumn{2}{c|}{KBP17}  \\
    \cline{2-3}\cline{4-5}\cline{6-7}
      & Macro & Micro & Macro& Micro & Macro& Micro \\
      \hline
    \multicolumn{7}{|c|}{Prototypical Network}\\
    \hline
    FS-Base\citep{snell2017prototypical}&66.2$\pm$3.8&63.8$\pm$4.1&44.1$\pm$1.6&44.0$\pm$2.3&67.1$\pm$1.7&68.0$\pm$1.6\\
    FS-Lexfree \citep{lu-etal-2019-distilling}&50.4$\pm$2.8&50.7$\pm$3.6&24.9$\pm$1.1&20.5$\pm$1.4&60.8$\pm$2.7&60.4$\pm$3.7\\
    FS-Cluster \citep{lai-etal-2020-extensively}&69.9$\pm$1.9&67.3$\pm$2.2&43.8$\pm$1.4&43.6$\pm$1.5&67.6$\pm$2.4&68.7$\pm$2.6\\
    FS-Causal (Ours)&\textbf{76.8}$\pm$0.6&\textbf{76.3}$\pm$0.7&\textbf{51.8}$\pm$0.5&\textbf{55.1}$\pm$0.4&\textbf{72.6}$\pm$0.9&\textbf{74.9}$\pm$0.9\\
    \hline
    \multicolumn{7}{|c|}{Relation Network}\\
    \hline
    FS-Base\citep{sung2018learning}&65.7$\pm$3.9&66.9$\pm$2.1&51.0$\pm$1.1&55.6$\pm$1.6&66.8$\pm$2.2&71.1$\pm$2.0\\
    FS-Lexfree \citep{lu-etal-2019-distilling}&59.1	$\pm$6.4&59.6$\pm$3.6&42.9$\pm$1.0&45.4$\pm$2.1&63.5$\pm$1.9&65.7$\pm$2.0\\
    FS-Cluster \citep{lai-etal-2020-extensively}&54.4$\pm$2.9&57.2$\pm$3.2&45.8$\pm$2.3&51.4$\pm$1.4&57.5$\pm$4.2&62.0$\pm$3.2\\
    FS-Causal (Ours)&\textbf{65.0}$\pm$2.1&\textbf{69.3}$\pm$1.5&\textbf{53.6}$\pm$0.7&\textbf{57.9}$\pm$1.0&\textbf{66.9}$\pm$0.1&\textbf{72.7}$\pm$0.9\\
    \hline
    \end{tabular}}
    \end{center}
    \caption{F1 score of 5-shot FSED on dev set. $\pm$ is the standard deviation of 5 random training rounds.}
    \label{tab:devresult}
  \end{table*}

  \begin{table}
    \centering
    \resizebox{0.45\textwidth}{!}{\begin{tabular}{lccc}
    \hline
    &ACE&MAVEN&KBP\\
    \hline
    Proto (support) &\textbf{76.25}&\textbf{55.11}&74.80\\
    Proto (query) & 65.75& 37.89& 69.25\\
    Proto (support + query) & 74.03&51.19&\textbf{74.93}\\
    \hline
    Relation (support) & 68.40 & 54.34&69.09\\
    Relation (query)&66.31&\textbf{57.85}&71.59\\
    Relation (support + query)&\textbf{69.31}&56.93&\textbf{72.65}\\
    \hline
    \end{tabular}}
    \caption{Micro F1 score of 5-shot FSED on dev set. (support) means the backdoor adjustment is used in the support set, (query) means the backdoor adjustment is used in the query.}
    \label{tab:ab1}
  \end{table}
   \begin{table}
    \centering
    \resizebox{0.40\textwidth}{!}{\begin{tabular}{lccc}
    \hline
    &ACE&MAVEN&KBP\\
    \hline
    Optimizer &AdamW&AdamW&AdamW\\
    Learning rate & 2e-5& 2e-5& 2e-5\\
    Warmup step & 40 & 240& 50\\
    Batch size&1&1&1\\
    patience & 15 & 15& 15\\
    max epoch num&80 & 80& 80\\
    batches per epoch &40 &240& 50\\
    $\lambda$ for $P(T|C)$&0.5&0.5&0.5\\
    FS-Causal (Prototypical)&S&S&S+Q\\
    FS-Causal (Relation)&S+Q&Q&S+Q\\
    \hline
    \end{tabular}}
    \caption{Hyperparameters of metric-based methods. For FS-Causal, S means the backdoor adjustment is used in the support set, Q means the backdoor adjustment is used in the query.}
    \label{tab:hyper}
  \end{table}
  \begin{table}
    \centering
    \resizebox{0.45\textwidth}{!}{\begin{tabular}{lccc}
    \hline
    &ACE&MAVEN&KBP\\
    \hline
    Optimizer &AdamW&AdamW&AdamW\\
    Learning rate (Pretrain) & 1e-5& 1e-5& 1e-5\\
    batches per epoch (Pretrain) &50 &200&200\\
    Warmup step (Pretrain) & 50 & 200&200\\
    Batch size (Pretrain)&128&128&128\\
    patience (Pretrain)& 10 & 10& 10\\
    max epoch num (Pretrain)&50 & 50&50\\
    Learning rate (Finetune) & 2e-5& 2e-5& 2e-5\\
    Learning rate (Finetune*) & 1e-3& 1e-3& 1e-3\\
    Finetuning Step (Finetune)& 20& 20& 20\\
    Finetuning Step (Pretrain + Finetune)& 10& 10& 10\\
    \hline
    \end{tabular}}
    \caption{Hyperparameters of finetuning-based methods.}
    \label{tab:hyper2}
  \end{table}
  From Appendix-A, we can obtain:
  $$
  \begin{aligned}
    &\sum_{t\in T}\sum_{s\in S} P(t|e) p(s|\mathcal{C},t)\\
    =&\sum_{t\in T}\sum_{s\in S} P(s,t|do(\mathcal{C}),e,q)\\
    =&1
  \end{aligned}
  $$
  
  \subsection{Prototypical Network} 
  For prototypical network, $g(\mathbf{r},\mathbf{q})=(\mathbf{r}-\mathbf{q})^2$. Let $\phi(s,q) = |\mathbf{r}_s- \mathbf{q}|$ and $f(x) = x^2,x>0$.
   $$
   \begin{aligned}
    &g(\sum_{t\in T}\sum_{s\in S} P(t|e) p(s|\mathcal{C},t)\mathbf{r}_s,\mathbf{q})\\
    =&[\sum_{t\in T}\sum_{s\in S} P(t|e) p(s|\mathcal{C},t) \mathbf{r}_s-\mathbf{q}]^2\\
    =&[\sum_{t\in T}\sum_{s\in S} P(t|e) p(s|\mathcal{C},t) \mathbf{r}_s\\
    &-\sum_{t\in T}\sum_{s\in S} P(t|e) p(s|\mathcal{C},t) \mathbf{q}]^2\\
    =&[\sum_{t\in T}\sum_{s\in S} P(t|e) P(s|\mathcal{C},t) |\mathbf{r}_s- \mathbf{q}|]^2\\
    =&f[\sum_{t\in T}\sum_{s\in S} P(t|e) P(s|\mathcal{C},t) \phi(s,q) ]\\
    \propto&f(P(Y|do(C=\mathcal{C}),E=e,Q=q))
   \end{aligned}
   $$

  \subsection{Relation Network}
  Let $g(\mathbf{r},\mathbf{q})=F[\mathbf{r} \oplus \mathbf{q} \oplus |\mathbf{r}-\mathbf{q}|]$. Let $\phi (s,q) = g(\mathbf{r}_s,\mathbf{q})$ and $f(x) = x$
  
   $$
   \begin{aligned}
    &g(\sum_{t\in T}\sum_{s\in S} P(t|e) P(s|\mathcal{C},t)\mathbf{r}_s,\mathbf{q})\\
    =&F[\sum_{t\in T}\sum_{s\in S} P(t |e) P(s|\mathcal{C},t)\mathbf{r}_s \oplus \mathbf{q}\\
    & \oplus |\sum_{t\in T}\sum_{s\in S}P(t |e) P(s|\mathcal{C},t)\mathbf{r}_s-\mathbf{q}|]\\
    =&F[\sum_{t\in T}\sum_{s\in S} P(t |e) P(s|\mathcal{C},t)\mathbf{r}_s  \\
    &\oplus \sum_{t\in T}\sum_{s\in S} P(t |e) P(s|\mathcal{C},t) \mathbf{q}\\
    & \oplus |\sum_{t\in T}\sum_{s\in S}P(t |e) P(s|\mathcal{C},t)\mathbf{r}_s\\
    &-\sum_{t\in T}\sum_{s\in S}P(t |e) P(s|\mathcal{C},t)\mathbf{q}|]\\
    \approx &F[\sum_{t\in T}\sum_{s\in S} P(t |e) P(s|\mathcal{C},t)\mathbf{r}_s  \\
    &\oplus \sum_{t\in T}\sum_{s\in S} P(t |e) P(s|\mathcal{C},t) \mathbf{q}\\
    & \oplus \sum_{t\in T}\sum_{s\in S}P(t |e) P(s|\mathcal{C},t)|\mathbf{r}_s-\mathbf{q}|]\\
    =&\sum_{t\in T}\sum_{s\in S} P(t|e) P(s|\mathcal{C},t) g(\mathbf{r},\mathbf{q})\\
    =&f[\sum_{t\in T}\sum_{s\in S} P(t|e) P(s|\mathcal{C},t)\phi(s,q)]\\
    \propto&f(P(Y|do(C=\mathcal{C}),E=e,Q=q))
    \\
   \end{aligned}
   $$

  Here, we assume that the feature representations of the same event type in support are close to each other so that $|\sum_sp_s\mathbf{r}_s-\sum_sp_sq|\approx \sum_sp_s|\mathbf{r}_s-q|$.
  
  \section{Implementation Details}
  \label{sec:detail}
  All of our experiments are implemented on one Nvidia TITAN RTX. Our implementation is based on HuggingFace’s Transformers \citep{wolf2019huggingface} and Allennlp \citep{gardner2018allennlp}. We tune the hyperparameters based on the dev performance. We train each model 5 times with different random seed, and when evaluating, we sample 4 different support sets. 
  
  \paragraph{Metric-based Methods} The hyperparameter is shown in Table \ref{tab:hyper}. During training, the support set and the query is sampled in training set, the query contains 2 positive instances and 10 negative instances (5 times of positive instances). During validating, the support set and the query is sampled in dev set, the query contains 10 positive instances and 100 negative instances (10 times of positive instances). The results of dev set are shown in Table \ref{tab:devresult}. For FS-Causal, we found that there is an impact on whether backdoor adjustment is applied separately to the support set and query, as shown in Table \ref{tab:ab1}. Based on the best results of the dev set, we evaluate it on the test set.

  \paragraph{Finetuning-based Methods} The hyperparameter is shown in Table \ref{tab:hyper2}. For pretraining, we train a supervised event detection model using the training set. For finetuning, we use the support set to finetune the parameters of the event detection model and then detect the event in query.

\end{document}